\documentclass[11pt,letterpaper]{article}
\usepackage[letterpaper]{geometry}
\usepackage{acl2012}
\usepackage{times}
\usepackage{latexsym}
\usepackage{amsmath}
\usepackage{multirow}
\usepackage{url}
\makeatletter
\newcommand{\@BIBLABEL}{\@emptybiblabel}
\newcommand{\@emptybiblabel}[1]{}
\makeatother
\usepackage[hidelinks]{hyperref}

\usepackage{caption}
\usepackage{algorithm}
\usepackage{algorithmicx}
\usepackage{algpseudocode}

\usepackage{tikz}
\usepackage{pgfplots}
\usepackage{array}
\usepackage{amsmath}
\usepackage{color}

\usepackage{arydshln}
\usepackage{graphicx}
\usepackage{caption}
\usepackage{subcaption}

\usepackage{pifont}
\usepackage{comment}

\usepackage{url}

\title{In-Order Transition-based Constituent Parsing}

\author{Jiangming Liu \and Yue Zhang\\
	    Singapore University of Technology and Design,\\
	    8 Somapah Road, Singapore, 487372\\
	    {\tt \{jiangming\_liu, yue\_zhang\}@sutd.edu.sg}}

\date{}

\begin{document}
\maketitle
\begin{abstract}
Both bottom-up and top-down strategies have been used for neural transition-based constituent parsing.
The parsing strategies differ in terms of the order in which they recognize productions in the derivation tree, where bottom-up strategies and top-down strategies take post-order and pre-order traversal over trees, respectively.
Bottom-up parsers benefit from rich features from readily built partial parses, but lack lookahead guidance in the parsing process; top-down parsers benefit from non-local guidance for local decisions, but rely on a strong encoder over the input to predict a constituent hierarchy before its construction.
To mitigate both issues, we propose a novel parsing system based on in-order traversal over syntactic trees, designing a set of transition actions to find a compromise between bottom-up constituent information and top-down lookahead information.
Based on stack-LSTM, our psycholinguistically motivated constituent parsing system achieves 91.8 F$_1$ on WSJ benchmark.
Furthermore, the system achieves 93.6 F$_1$ with supervised reranking and 94.2 F$_1$ with semi-supervised reranking, which are the best results on the WSJ benchmark.
\end{abstract}

\section{Introduction}
Transition-based constituent parsing employs sequences of local transition actions to construct constituent trees over sentences.
There are two popular transition-based constituent parsing systems, namely bottom-up parsing \cite{sagae:2005,zhang:2009,zhu:2013,watanabe:2015} and top-down parsing \cite{dyer:2016,kuncoro:2017}.
The parsing strategies differ in terms of the order in which they recognize productions in the derivation tree.

\begin{figure}
\begin{center}
\includegraphics[width=8 cm,height=10cm]{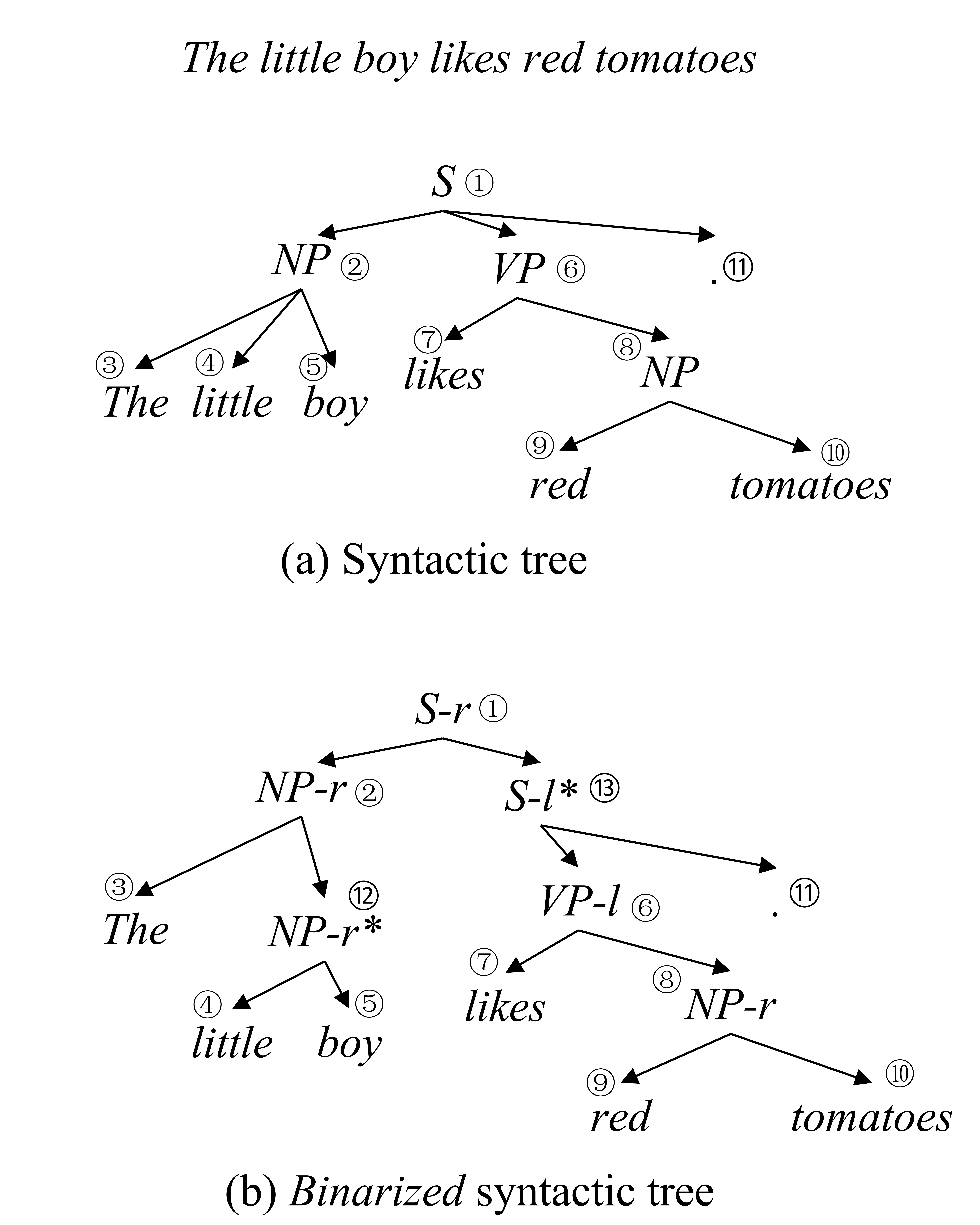}
\end{center}
\caption{\label{trees} Syntactic trees of the sentence ``The little boy likes red tomatoes.". (a) syntactic tree; (b) binarized syntactic tree, where $r$ and $l$ mean the head is the right branch and the left branch, respectively, and $*$ means this constituent is not completed.}
\end{figure}

The process of bottom-up parsing can be regarded as \textit{post-order} traversal over a constituent tree.
For example, given the sentence in Figure \ref{trees}, a bottom-up shift-reduce parser takes the action sequence in Table \ref{nodetrace}(a)\footnote{The action sequence is taken on unbinarized trees.} to build the output, where the word sequence ``The little boy" is first read, and then an \textit{NP} recognized for the word sequence.
After the system reads the verb ``likes" and an its subsequent \textit{NP}, a \textit{VP} is recognized.
The full order of recognition for the tree nodes is \textcircled{\scriptsize{3}}$\rightarrow$\textcircled{\scriptsize{4}}$\rightarrow$\textcircled{\scriptsize{5}}$\rightarrow$\textcircled{\scriptsize{2}}$\rightarrow$\textcircled{\scriptsize{7}}$\rightarrow$\textcircled{\scriptsize{9}}$\rightarrow$\textcircled{\scriptsize{10}}$\rightarrow$\textcircled{\scriptsize{8}}$\rightarrow$\textcircled{\scriptsize{6}}$\rightarrow$\textcircled{\scriptsize{11}}$\rightarrow$ \textcircled{\scriptsize{1}}.
When making local decisions, rich information is available from readily built partial trees \cite{zhu:2013,watanabe:2015,cross:2016}, which contributes to local disambiguation.
However, there is lack of top-down guidance from lookahead information, which can be useful \cite{johnson:1998,roark:1999,charniak:2000,liu:2017}.
In addition, \textit{binarization} must be applied to trees, as shown in Figure \ref{trees}(b), to ensure a constant number of actions \cite{sagae:2005}, and to take advantage of lexical head information \cite{collins:2003}.
However, such \textit{binarization} requires a set of language-specific rules, which hampers adaptation of parsing to other languages.

On the other hand, the process of top-down parsing can be regarded as \textit{pre-order} traversal over a tree.
Given the sentence in Figure \ref{trees}, a top-down shift-reduce parser takes the action sequence in Table \ref{nodetrace}(b) to build the output, where an \textit{S} is first made and then an \textit{NP} is generated.
After that, the system makes decision to read the word sequence ``The little boy" to complete the \textit{NP}.
The full order of recognition for the tree nodes is \textcircled{\scriptsize{1}}$\rightarrow$\textcircled{\scriptsize{2}}$\rightarrow$\textcircled{\scriptsize{3}}$\rightarrow$\textcircled{\scriptsize{4}}$\rightarrow$\textcircled{\scriptsize{5}}$\rightarrow$\textcircled{\scriptsize{6}}$\rightarrow$\textcircled{\scriptsize{7}}$\rightarrow$\textcircled{\scriptsize{8}}$\rightarrow$\textcircled{\scriptsize{9}}$\rightarrow$\textcircled{\scriptsize{10}}$\rightarrow$ \textcircled{\scriptsize{11}}.
The top-down lookahead guidance contributes to non-local disambiguation.
However, it is difficult to generate a constituent before its sub constituents have been realized, since no explicit features can be extracted from their subtree structures.
Thanks to the use of recurrent neural networks, which makes it possible to represent a sentence globally before syntactic tree construction, seminal work of neural top-down parsing directly generates bracketed constituent trees using sequence-to-sequence models \cite{vinyals:2015}.
\newcite{dyer:2016} design set of top-down transition actions for standard transition-based parsing.

In this paper, we propose a novel transition system for constituent parsing, mitigating issues of both bottom-up and top-down systems by finding a compromise between bottom-up constituent information and top-down lookahead information.
The process of the proposed constituent parsing can be regarded as \textit{in-order} traversal over a tree.
Given the sentence in Figure \ref{trees}, the system takes the action sequence in Table \ref{nodetrace}(c) to build the output.
The system reads the word ``The" and then projects an NP, which is based on bottom-up evidence.
\begin{figure}[!tp]
\centering
\renewcommand{\arraystretch}{0.8}
\begin{tabular}{>{\small}r>{\small}l<{\hspace{-6pt}}>{\small}c<{\hspace{-6pt}}>{\small}c}
\hline
stack & buffer & action & node \\
\hline
{[]} & {[The little ...]} & \textsc{Shift} & \textcircled{\scriptsize{3}} \\
{[The]} & {[little boy ...]} & \textsc{Shift} & \textcircled{\scriptsize{4}} \\
{[The little]} & {[boy likes ...]} & \textsc{Shift} & \textcircled{\scriptsize{5}} \\
{[... little boy]} & {[likes red ...]} & \textsc{Reduce-r-NP} & \textcircled{\scriptsize{2}} \\ 
... & ... & ... & ... \\
\hline
\\
\multicolumn{4}{c}{(a) bottom-up system} \\
\\
\hline
stack & buffer & action & node \\
\hline
{[]} & {[The little ...]} & \textsc{NT-S} & \textcircled{\scriptsize{1}} \\
{[(S]} & {[The little ...]} & \textsc{NT-NP} & \textcircled{\scriptsize{2}} \\
{[(S (NP]} & {[The little ...]} & \textsc{Shift} & \textcircled{\scriptsize{3}} \\
{[... (NP The]} & {[little boy ...]} & \textsc{Shift} & \textcircled{\scriptsize{4}} \\
{[... The little]} & {[boy likes ...]} & \textsc{Shift} & \textcircled{\scriptsize{5}} \\
{[... little boy]} & {[likes red ...]} & \textsc{Reduce} & / \\ 
... & ... & ... & ... \\
\hline
\\
\multicolumn{4}{c}{(b) top-down system} \\
\\
\hline
stack & buffer & action & node \\
\hline
{[]} & {[The little ...]} & \textsc{Shift} & \textcircled{\scriptsize{3}} \\
{[The]} & {[little boy ...]} & \textsc{PJ-NP} & \textcircled{\scriptsize{2}} \\
{[The NP]} & {[little boy ...]} & \textsc{Shift} & \textcircled{\scriptsize{4}} \\
{[... NP little]} & {[boy likes...]} & \textsc{Shift} & \textcircled{\scriptsize{5}} \\
{[... little boy]} & {[likes red ...]} & \textsc{Reduce} & / \\ 
... & ... & ... & ... \\
\hline
\\
\multicolumn{4}{c}{(c) in-order system} \\
\end{tabular}
\caption{Action sequences of three types of transition constituent parsing system (till the recognition of ``(NP The little boy)"). Details of the action system are introduced in Section 2.1, Section 2.2 and Section 3, respectively. }
\label{nodetrace}
\end{figure}
After this, based on the projected NP, the system reads the word sequence ``little boy", with top-down guidance from NP.
Similarly, based on the completed constituent ``(NP The little boy)", the system projects an S, with the bottom-up evidence.
With the S and the word ``likes", the system projects an VP, which can serve as top-down guidance.
The full order of recognition for the tree nodes is \textcircled{\scriptsize{3}}$\rightarrow$\textcircled{\scriptsize{2}}$\rightarrow$\textcircled{\scriptsize{4}}$\rightarrow$\textcircled{\scriptsize{5}}$\rightarrow$\textcircled{\scriptsize{1}}$\rightarrow$\textcircled{\scriptsize{7}}$\rightarrow$\textcircled{\scriptsize{6}}$\rightarrow$\textcircled{\scriptsize{9}}$\rightarrow$\textcircled{\scriptsize{8}}$\rightarrow$\textcircled{\scriptsize{10}}$\rightarrow$ \textcircled{\scriptsize{11}}.
Compared to \textit{post-order} traversal, \textit{in-order} traversal can potentially resolve non-local ambiguity better by top-down guidance.
Compared to \textit{pre-order} traversal, \textit{in-order} traversal can potentially resolve local ambiguity better by bottom-up evidence.

Furthermore, \textit{in-order} traversal is psycho-linguistically motivated \cite{roark:2009,steedman:2000}.
Empirically, a human reader comprehends sentences by giving lookahead guesses for constituents.
For example, when reading a word ``likes", a human reader could guess that it could be a start of a constituent VP, instead of waiting to read the object ``red tomatoes", which is the procedure of a bottom-up system.

We compare our system with the two baseline systems (i.e. a top-down system and a bottom-up system) under the same neural transition-based framework of \newcite{dyer:2016}. 
Our final models outperform both of the bottom-up and top-down transition-based constituent parsing by achieving a 91.8 $F_1$ in English and a 86.1 $F_1$ in Chinese for greedy fully-supervised parsing, respectively.
Furthermore, our final model obtains a 93.6 F$_1$ with supervised reranking \cite{choe:2016} and a 94.2 F$_1$ with semi-supervised reranking, achieving the state-of-the-art results on constituent parsing on English benchmark.
By converting to Stanford dependencies, our final model achieves the state-of-the-art results on dependency parsing by obtaining a 96.2\% UAS and a 95.2\% LAS.
To our knowledge, we are the first to systematically compare top-down and bottom-up constituent parsing under the same neural framework.
We release our code at \url{https://github.com/LeonCrashCode/InOrderParser}.

\section{Transition-based constituent parsing}
Transition-based constituent parsing takes a left-to-right scan of the input sentence, where a stack is used to maintain partially constructed phrase-structures, while the input words are stored in a buffer.
Formally, a \textit{state} is defined as $[\sigma, i, f]$, where $\sigma$ is the stack, $i$ is the front index of the buffer, and $f$ is a bool value showing if the parsing is finished.
At each step, a transition action is applied to consume an input word or construct a new phrase-structure.
Different parsing systems employ their own sets of actions.

\subsection{Bottom-up system}
We take the bottom-up system of \newcite{sagae:2005} as our bottom-up baseline.
Given a state, the set of transition actions are
\begin{itemize}
\item \textsc{Shift}: pop the front word from the buffer, and push it onto the stack.
\item \textsc{Reduce-l/r-X}: pop the top two constituents off the stack, combine them into a new constituent with label X, and push the new constituent onto the stack.
\item \textsc{Unary-X}: pop the top constituent off the stack, raise it to a new constituent with label X, and push the new constituent onto the stack.
\item \textsc{Finish}: pop the root node off the stack and ends parsing.
\end{itemize}
The bottom-up parser can be summarized as the deductive system in Figure \ref{deduction}(a).
Given the sentence with the binarized syntactic tree in Figure \ref{trees}(b), the sequence of actions \textsc{Shift}, \textsc{Shift}, \textsc{Shift}, \textsc{Reduce-r-NP}, \textsc{Reduce-r-NP}, \textsc{Shift}, \textsc{Shift}, \textsc{Shift}, \textsc{Reduce-r-NP}, \textsc{Reduce-l-VP}, \textsc{Shift}, \textsc{Reduce-l-S}, \textsc{Reduce-r-S} and \textsc{Finish}, can be used to construct its constituent tree.

\subsection{Top-down system}
We take the top-down system of \newcite{dyer:2016} as our top-down baseline.
Given a state, the set of transition actions are
\begin{itemize}
\item \textsc{Shift}: pop the front word from the buffer, and push it onto the stack.
\item \textsc{NT-X}: open a nonterminal with label X on top of the stack.
\item \textsc{Reduce}: repeatedly pop completed subtrees or terminal symbols from the stack until an open nonterminal is encountered, and then this open NT is popped and used as the label of a new constituent that has the popped subtrees as its children. This new completed constituent is pushed onto the stack as a single composite item.
\end{itemize}
\begin{figure}[!tp]
\centering
\renewcommand{\arraystretch}{0.8}
\begin{tabular}{cc}
\multirow{2}{*}{\textsc{Shift}} & $[\sigma, i, false]$ \\
\cline{2-2}
&$[\sigma|w_i, i+1, false]$ \\
\\
\multirow{2}{*}{\textsc{Reduce-l/r-X}} & $[\sigma|s_1|s_0, i, false]$ \\
\cline{2-2}
&$[\sigma|X_{s_1s_0}, i, false]$\\
\\
\multirow{2}{*}{Unary-X} & $[\sigma|s_0, i, false]$ \\
\cline{2-2} 
&$[\sigma|X_{s_0}, i, false]$\\
\\
\multirow{2}{*}{\textsc{Finish}} & $[\sigma, i, false]$ \\
\cline{2-2}
&$[\sigma, i, true]$\\
\\
\multicolumn{2}{c}{(a) bottom-up system} \\
\\
\multirow{2}{*}{\textsc{Shift}} & $[\sigma, i, /]$ \\
\cline{2-2}
&$[\sigma|w_i, i+1, /]$ \\
\\
\multirow{2}{*}{\textsc{NT-X}} & $[\sigma, i, /]$ \\
\cline{2-2}
&$(\sigma|X, i, /]$\\
\\
\multirow{2}{*}{\textsc{Reduce}} & $[\sigma|X|s_j|...|s_0, i, /]$ \\
\cline{2-2} 
&$[\sigma|X_{s_j...s_0}, i, /]$\\
\\
\multicolumn{2}{c}{(b) top-down system} \\
\\
\multirow{2}{*}{\textsc{Shift}} & $[\sigma, i, false]$ \\
\cline{2-2}
&$[\sigma|w_i, i+1, false]$ \\
\\
\multirow{2}{*}{\textsc{PJ-X}} & $[\sigma|s_0, i, false]$ \\
\cline{2-2}
&$(\sigma|s_0|X, i, false]$\\
\\
\multirow{2}{*}{\textsc{Reduce}} & $[\sigma|s_j|X|s_{j-1}|...|s_0, i, false]$ \\
\cline{2-2} 
&$[\sigma|X_{s_js_{j-1}...s_0}, i, false]$\\
\\
\multirow{2}{*}{\textsc{Finish}} & $[\sigma, i, false]$ \\
\cline{2-2} 
&$[\sigma, i, true]$\\
\\
\multicolumn{2}{c}{(c) in-order system} \\
\end{tabular}
\caption{Different transition systems. The start state is $[\phi, 0, false]$ and the final state is $[\sigma, n, true]$.}
\label{deduction}
\end{figure}
The deduction system for the process is shown in Figure \ref{deduction}(b)\footnote{Due to unary decision, we use completed marks to make finish decision, except for top-down system.}.
Given the sentence in Figure \ref{trees}, the sequence of actions \textsc{NT-S}, \textsc{NT-NP}, \textsc{Shift}, \textsc{Shift}, \textsc{Shift}, \textsc{Reduce}, \textsc{NT-VP}, \textsc{Shift}, \textsc{NT-NP}, \textsc{Shift}, \textsc{Shift}, \textsc{Reduce}, \textsc{Reduce}, \textsc{Shift} and \textsc{Reduce}, can be used to construct its constituent tree.

\section{In-order system}
We propose a novel in-order system for transition-based constituent parsing.
Similar to the bottom-up and top-down systems, the in-order system maintains a stack and a buffer for representing a state.
The set of transition actions are defined as:
\begin{itemize}
\item \textsc{Shift}: pop the front word from the buffer, and push it onto the stack.
\item \textsc{PJ-X}: project a nonterminal with label X on top of the stack.
\item \textsc{Reduce}: repeatedly pop completed subtrees or terminal symbols from the stack until an projected nonterminal is encountered, and then this projected nonterminal is popped and used as the label of a new constituent, and furthermore, one more item on the top of stack is popped as the leftmost child of the new constituent and the popped subtrees as its rest children. This new completed constituent is pushed onto the stack as a single composite item.
\item \textsc{Finish}: pop the root node off the stack and ends parsing.
\end{itemize}
The deduction system for the process is shown in Figure \ref{deduction}(c).
Given the sentence in Figure \ref{trees}, the sequence of actions \textsc{Shift}, \textsc{PJ-NP}, \textsc{Shift}, \textsc{Shift}, \textsc{Reduce}, \textsc{PJ-S}, \textsc{Shift}, \textsc{PJ-VP}, \textsc{Shift}, \textsc{PJ-NP}, \textsc{Shift}, \textsc{Reduce}, \textsc{Reduce}, \textsc{Shift}, \textsc{Reduce}, \textsc{Finish} can be used to construct its constituent tree.

\textbf{Variants}~~~~
The in-order system can be generalized into variants by modifying $k$, the number of leftmost nodes traced before the parent node.
For example, given the tree ``(S a b c d)", the traversal is ``a S b c d" if $k=1$ while the traversal is ``a b S c d" if $k=2$.
We name each variant with a certain $k$ value as $k$-in-order systems.
In this paper, we only investigate the in-order system with $k=1$, the 1-in-order system.
Note that the top-down parser can be regarded as a special case of a generalized version of the in-order parser with $k=0$, and the bottom-up parser can be regarded as a special case with $k=\infty$.
\begin{figure}
\begin{center}
\includegraphics[width=8 cm,height=6.85cm]{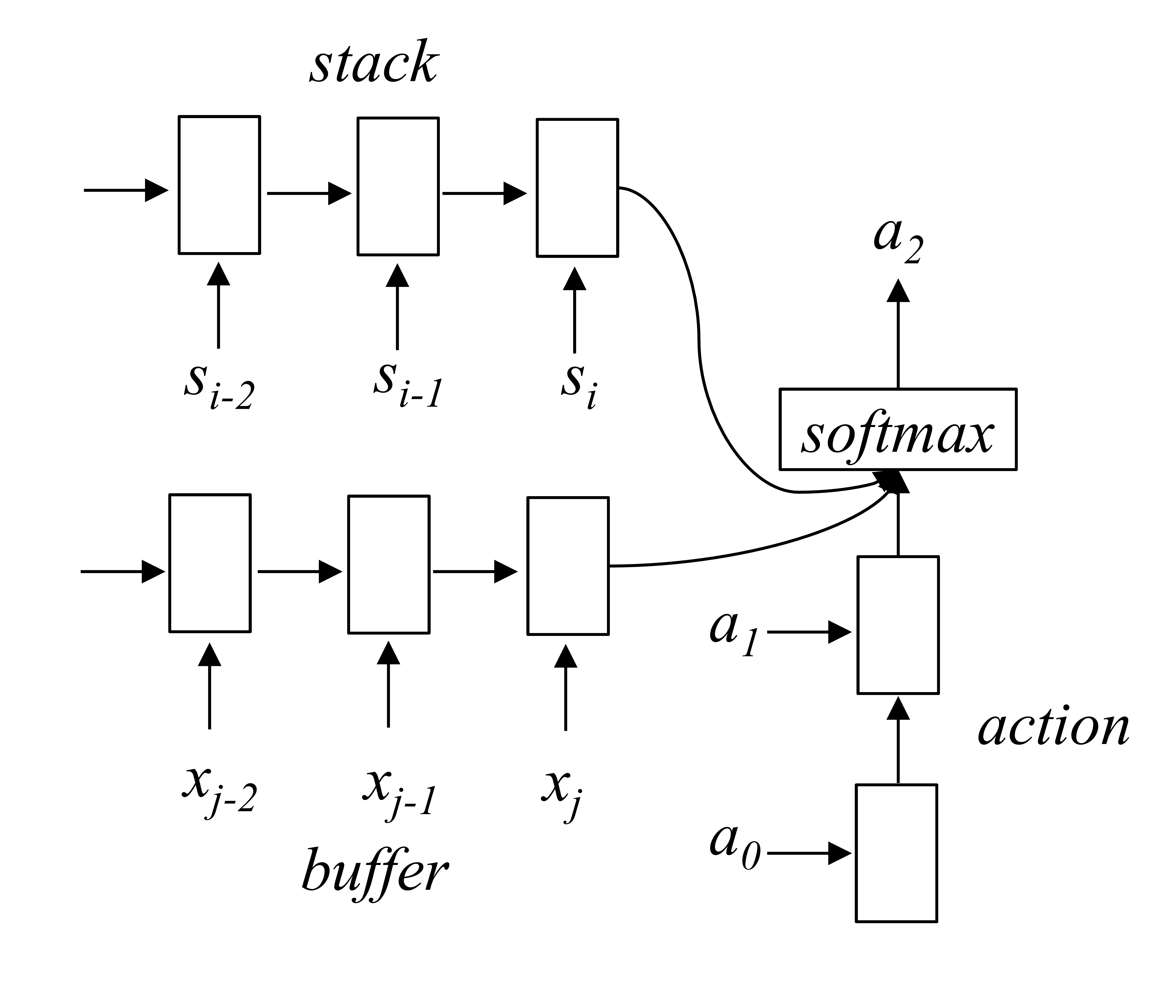}
\end{center}
\caption{\label{stacklstm} Framework of our transition-based parsers.}
\end{figure}
\section{Neural parsing model}
We employ the stack-LSTM parsing model of \newcite{dyer:2016} for the three types of transition-based parsing systems in Section 2.1, 2.2 and 3, respectively, where a stack-LSTM is used to represent the stack, a stack-LSTM is used to represent the buffer, and a vanilla LSTM is used to represent the action history, as shown in Figure \ref{stacklstm}.

\subsection{Word representation}
We follow \newcite{dyer:2015}, representing each word using three different types of embeddings, including pretrained word embedding, $\overline{e}_{w_i}$, which is not fine-tuned during training of the parser, randomly initialized embeddings $e_{w_i}$, which is fine-tuned, and the randomly initialized part-of-speech embeddings, which is fine-tuned.
The three embeddings are concatenated, and then fed to nonlinear layer to derive the final word embedding: 
\begin{equation*}
x_i = f(W_{input}[e_{p_i}; \overline{e}_{w_i}; e_{w_i}]+b_{input}),
\end{equation*}
where $W_{input}$ and $b_{input}$ are model parameters, $w_i$ and $p_i$ denote the form and the POS tag of the $i$th input word, respectively, and $f$ is an nonlinear function.
In this paper, we use ReLu for $f$. 

\subsection{Stack representation}
\begin{figure}
\begin{center}
\includegraphics[width=8 cm,height=9cm]{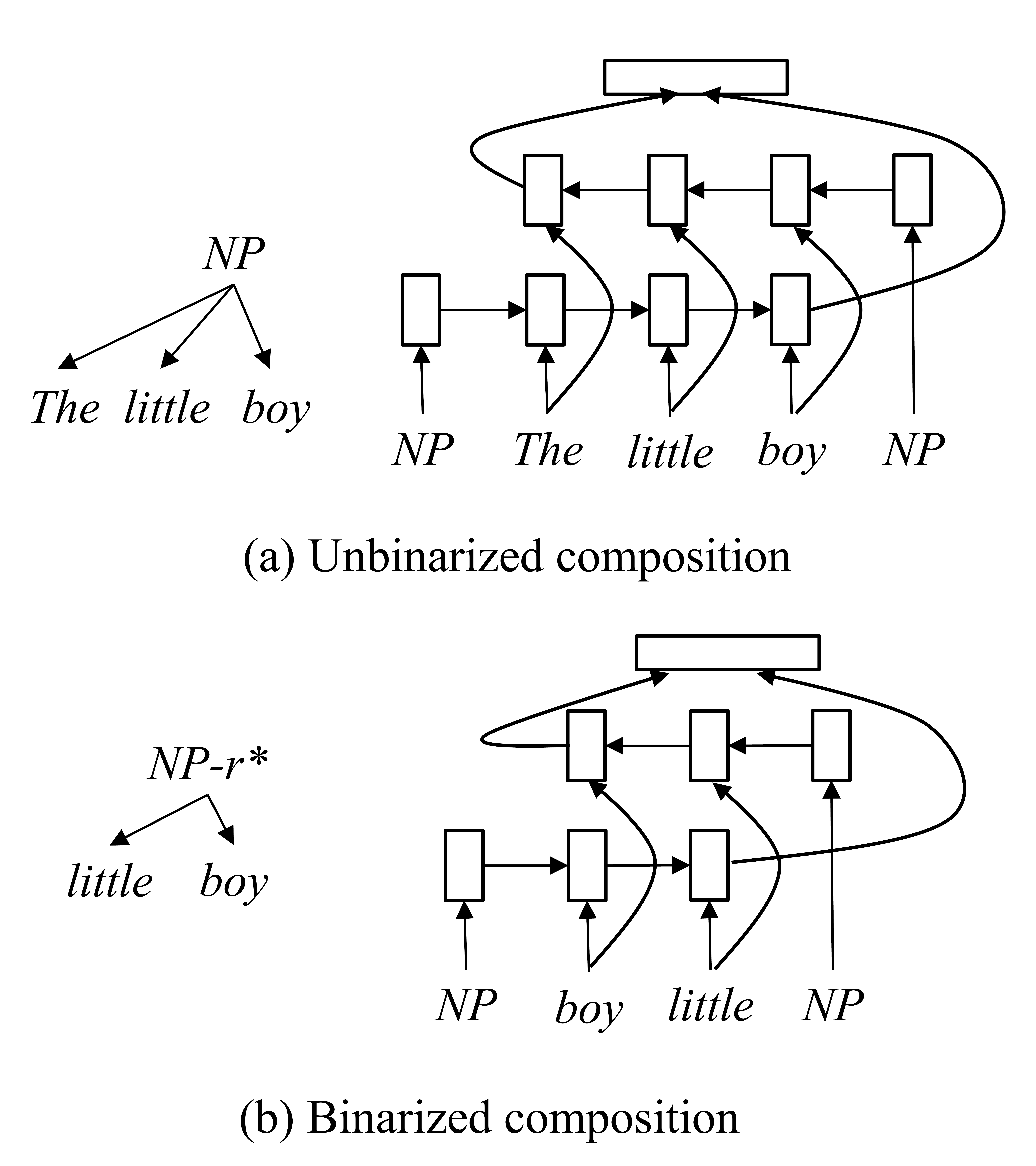}
\end{center}
\caption{\label{composition} The composition function. (a) is for unbinarized trees and (b) is for binarized trees, where ``NP-r*" means that ``little boy" is a non-completed noun phrase with head ``boy".}
\end{figure}

We employ a bidirectional LSTM as the composition function to represent constituents on stack\footnote{To be fair, we use bidirectional LSTM as composition function for all parsing systems}.
For top-down parsing and in-order parsing, following \newcite{dyer:2016}, as shown in Figure \ref{composition}(a), the composition representation $s_{comp}$ is computed as:
\begin{equation*}
\begin{array}{cl}
s_{comp} = & (\text{LSTM}_{fwd}[e_{nt}, s_{0}, ..., s_{m}]; \\
&\text{LSTM}_{bwd}[e_{nt}, s_{m}, ..., s_{0}] ), \\
\end{array}
\end{equation*}
where $e_{nt}$ is the representation of a non-terminal, $s_{j}, j\in[0,m]$ is the $j$th child node, and $m$ is the number of the child nodes. 
For bottom-up parsing, we make use of head information in the composition function by requiring the order that the head node is always before the non-head node in the bidirectional LSTM, as shown in Figure \ref{composition}(b)\footnote{A bidirectional LSTM consists of two LSTMs, making it balanced for composition. However, they have different parameters so that one represents information of head-first while other represents information of head-last.}.
The binarized composition is computed as:
\begin{equation*}
\begin{array}{cl}
s_{bcomp} = & (\text{LSTM}_{fwd}[e_{nt}, s_{h}, s_{o}]; \\
&\text{LSTM}_{bwd}[e_{nt}, s_{o}, s_{h}] ), \\
\end{array}
\end{equation*}
where $s_{h}$ and $s_{o}$ is the representation of head and non-head node, respectively. 

\subsection{Greedy action classification}
Given a sentence $w_0, w_1, ..., w_{n-1}$, where $w_i$ is the $i$th word, and $n$ is the length of the sentence, our parser makes local action classification decisions incrementally.
For the $k$th parsing state like [$s_j, ..., s_1, s_0$, $i$, \textit{false}], the probability distribution of the current action $p$ is:
\begin{equation*}
p = \textsc{SoftMax}(W[h_{stk}; h_{buf}; h_{ah}]+b)\tag{*},
\end{equation*}
where $W$ and $b$ are model parameters, the representation of stack information $h_{stk}$ is
\begin{equation*}
h_{stk} = \text{stack-LSTM}[s_0, s_1, ..., s_j],
\end{equation*}
the representation of buffer information $h_{buf}$ is
\begin{equation*}
h_{buf} = \text{stack-LSTM}[x_{i}, x_{i+1}, ..., x_{n}],
\end{equation*}
$x$ is the word representation, and the representation of action history $h_{ah}$ is
\begin{equation*}
h_{ah} = \text{LSTM}[e_{act_{k-1}}, e_{act_{k-2}}, ..., e_{act_0}],
\end{equation*}
where $e_{act_{k-1}}$ is the representation of action in $k$-1th parsing state.

\textbf{Training} ~~~
Our models are trained to minimize a cross-entropy loss objective with an $l_2$ regularization term, defined by
\begin{equation*}
L(\theta)=-\sum_i{\sum_j{log~p_{a_{ij}}}} + \frac{\lambda}{2}||\theta||^2,
\end{equation*}
where $\theta$ is the set of parameters, $p_{a_{ij}}$ is the probability of the $j$th action in the $i$th training example given by the model and $\lambda$ is a regularization hyper-parameter ($\lambda = 10^{-6}$).
We use stochastic gradient descent with 0.1 initialized learning rate with 0.05 learning rate decay.

\section{Experiments}
\subsection{Data}
We empirically compare our bottom-up, top-down and in-order parsers.
The experiments are carried on both English and Chinese.
For English data, we use the standard benchmark of WSJ sections in PTB \cite{Marcus:1993}, where the sections 2-21 are taken for training data, section 22 for development data and section 23 for test for both dependency parsing and constituency parsing.
We adopt the pretrained English word embeddings generated on the AFP portion of English Gigaword.

For Chinese data, we use the version 5.1 of the Penn Chinese Treebank (CTB) \cite{Xue:2005}.
We use articles 001- 270 and 440-1151 for training, articles 301-325 for system development, and articles 271-300 for final performance evaluation.
We adopt the pretrained Chinese word embeddings generated on the complete Chinese Gigaword corpus,

The POS tags in both the English data and the Chinese data are automatically assigned as the same as the work of \newcite{dyer:2016}, using Stanford tagger.
We follow the work of \newcite{choe:2016} and adopt the AFP portion of English Gigaword as the extra resources for the semi-supervised reranking.

\subsection{Settings}
\indent \textbf{Hyper-parameters}~~~
For both English and Chinese experiments, we use the same hyper-parameters as the work of \newcite{dyer:2016} without further optimization, as shown in Table \ref{parameter}.
\begin{table}[!tp]
\begin{center}
\renewcommand{\arraystretch}{0.8}
\begin{tabular}{>{\small}l|>{\small}c}
\hline
Parameter & Value \\
\hline
\hline
LSTM layer & 2 \\
Word embedding dim & 32 \\
English pretrained word embedding dim & 100 \\
Chinese pretrained word embedding dim & 80 \\
POS tag embedding dim & 12 \\
Action embedding dim & 16 \\
Stack-LSTM input dim & 128 \\
Stack-LSTM hidden dim & 128 \\
\hline
\end{tabular}
\end{center}
\caption{\label{parameter} Hyper-parameters.}
\end{table}

\noindent\textbf{Reranking experiments}~~~
Following the same reranking setting of \newcite{dyer:2016} and \newcite{choe:2016}, we obtain 100 samples from our bottom-up, top-down, and in-order model (section 4), respectively, with exponentiation strategy ($\alpha=0.8$) by using probability distribution of action (equation *).
We adopt the reranker of \newcite{choe:2016} as both our English fully-supervised reranker and semi-supervised reranker, and the generative reranker of \newcite{dyer:2016} as our Chinese supervised reranker.

\subsection{Development experiments}
Table \ref{dev} shows the development results of the three parsing systems.
The bottom-up system performs slightly better than the top-down system.
The in-order system outperforms both the bottom-up and the top-down system.
\begin{table}[!tp]
\begin{center}
\renewcommand{\arraystretch}{0.8}
\begin{tabular}{>{\small}l|>{\small}c|>{\small}c|>{\small}c}
\hline
Model &  LR & LP & F$_1$ \\
\hline
\hline
Top-down parser & 91.59 & 91.66 & 91.62\\
Bottom-up parser & 91.89 & 91.83 & 91.86\\
In-order parser & 91.98 & 91.86 & 91.92\\
\hline 
\end{tabular}
\end{center}
\caption{\label{dev}  Development results (\%) on WSJ 22.}
\end{table}

\subsection{Results}
Table \ref{test_en} shows the parsing results on the English test dataset.
We find that the bottom-up parser and the top-down parser have similar results under the greedy setting, and the in-order parser outperforms both of them.
Also, with supervised reranking, the in-order parser achieves the best results.
\begin{table}[!tp]
\begin{center}
\renewcommand{\arraystretch}{0.8}
\begin{tabular}{>{\small}l|>{\small}c}
\hline
Model &  F$_1$ \\
\hline
\hline
\multicolumn{2}{>{\small}l}{fully-supervision} \\
\hline
Top-down parser & 91.2 \\
Bottom-up parser & 91.3 \\
In-order parser & 91.8 \\
\hline
\multicolumn{2}{>{\small}l}{rerank} \\
\hline
Top-down parser & 93.3 \\ 
Bottom-up parser & 93.3 \\
In-order parser & 93.6 \\
\hline
\end{tabular}
\end{center}
\caption{\label{test_en}  Final results (\%) on WSJ section 23.}
\end{table}
\begin{table}[!tp]
\begin{center}
\renewcommand{\arraystretch}{0.8}
\begin{tabular}{>{\small}l|>{\small}c}
\hline
Model &  F$_1$ \\
\hline
\hline
\multicolumn{2}{>{\small}l}{fully-supervision} \\
\hline
\newcite{socher:2013} & 90.4 \\
\newcite{zhu:2013} & 90.4 \\
\newcite{vinyals:2015} & 90.7 \\
\newcite{watanabe:2015} & 90.7 \\
\newcite{shindo:2012} & 91.1 \\
\newcite{durrett:2015} & 91.1\\
\newcite{dyer:2016}  & 91.2\\
\newcite{cross:2016} & 91.3 \\
\newcite{liu:2017} & 91.7 \\
\hdashline
Top-down parser & 91.2 \\
Bottom-up parser & 91.3 \\
In-order parser & \textbf{91.8} \\
\hline
\multicolumn{2}{>{\small}l}{reranking} \\
\hline
\newcite{huang:2008} & 91.7 \\
\newcite{charniak:2005} & 91.5 \\
\newcite{choe:2016} & 92.6\\
\newcite{dyer:2016} & 93.3 \\
\newcite{kuncoro:2017} & \textbf{93.6} \\
\hdashline
Top-down parser & 93.3 \\
Bottom-up parser & 93.3 \\
In-order parser & \textbf{93.6} \\
\hline
\multicolumn{2}{>{\small}l}{semi-supervised reranking}\\
\hline
\newcite{choe:2016} & 93.8\\
In-order parser & \textbf{94.2} \\
\hline
\end{tabular}
\end{center}
\caption{\label{final_en_con}  Final results (\%) on WSJ section 23.}
\end{table}

\textbf{English constituent results}~~~We compare our models with previous work, as shown in Table \ref{final_en_con}.
With the fully-supervision setting\footnote{Here, we only consider the work of single model}, the in-order parser outperforms the state-of-the-art discrete parser \cite{shindo:2012,zhu:2013}, the state-of-the-art neural parsers \cite{cross:2016,watanabe:2015} and the state-of-the-art hybrid parsers \cite{durrett:2015,liu:2017}, achieving the state-of-the-art results.
With the reranking setting, the in-order parser outperforms the best discrete parser \cite{huang:2008} and have the same performance of \newcite{kuncoro:2017}, which extend the work of \newcite{dyer:2016} by adding gated attention mechanism on composition functions.
With the semi-supervised setting, the in-order parser outperforms the best semi-supervised parser \cite{choe:2016} by achieving 94.2 F$_1$.

\textbf{English dependency results}~~~As shown in Table \ref{final_en_dep}, by converting to Stanford Dependencies, without additional training data, our models achieves similar performance with the state-of-the-art system \cite{choe:2016}; with the same additional training data, our models achieves new state-of-the-art results on dependency parsing by achieving 96.2\% UAS and 95.2\% LAS on standard benchmark.
\begin{table}[!tp]
\begin{center}
\renewcommand{\arraystretch}{0.8}
\begin{tabular}{>{\small}l|>{\small}c>{\small}c}
\hline
Model &  UAS & LAS \\
\hline
\hline
\newcite{kiperwasser:2016}$\dagger$ & 93.9 & 91.9 \\
\newcite{cheng:2016} $\dagger$ & 94.1 & 91.5 \\
\newcite{andor:2016} & 94.6 & 92.8 \\
\newcite{dyer:2016} -re& 95.6 & 94.4 \\
\newcite{dozat:2017}$\dagger$ & 95.7 & 94.0 \\
\newcite{kuncoro:2017} -re &95.7 & 94.5 \\
\newcite{choe:2016} -sre & 95.9 & 94.1\\
\hdashline
In-order parser & 94.5 & 93.4 \\
In-order parser -re & 95.9 & 94.9 \\
In-order parser -sre & \textbf{96.2} & \textbf{95.2} \\
\hline
\end{tabular}
\end{center}
\caption{\label{final_en_dep}  Stanford Dependency accuracy (\%) on WSJ section 23. $\dagger$ means graph-based parsing. ``-re" means fully-supervised reranking and ``-sre" means semi-supervised reranking.}
\end{table}

\textbf{Chinese constituent results}~~~
Table \ref{final_ch} shows the final results on Chinese test dataset.
The in-order parser achieves the best results under hte fully-supervised setting.
With the supervised reranking, the in-order parser outperforms the state-of-the-art models by achieving 88.0 F$_1$. 

\textbf{Chinese dependency results}~~~
As shown in Table \ref{final_dep_ch}, by converting the results to dependencies\footnote{The Penn2Malt tool is used with Chinese head rules {https://stp.lingfil.uu.se/~nivre/research/Penn2Malt.html}}, our final model achieves the best results among transition-based parsing, and obtains comparable results to the state-of-the-art graph-based models.
\begin{table}[!tp]
\begin{center}
\renewcommand{\arraystretch}{0.8}
\begin{tabular}{>{\small}l|>{\small}c}
\hline
Parser &  F$_1$ \\
\hline
\hline
\multicolumn{2}{>{\small}l}{fully-supervision} \\
\hline
\newcite{zhu:2013}& 83.2 \\
\newcite{wang:2015} & 83.2 \\
\newcite{dyer:2016} & 84.6 \\
\newcite{liu:2017}& 85.5 \\
\hdashline
Top-down parser & 84.6 \\
Bottom-up parser & 85.7 \\
In-order parser & \textbf{86.1} \\
\hline
\multicolumn{2}{>{\small}l}{rerank} \\
\hline
\newcite{charniak:2005} & 82.3 \\
\newcite{dyer:2016} & 86.9 \\
\hdashline
Top-down parser & 86.9 \\
Bottom-up parser & 87.5 \\
In-order parser & \textbf{88.0} \\
\hline
\multicolumn{2}{>{\small}l}{semi-supervision}\\
\hline
\newcite{zhu:2013} & 85.6 \\
\newcite{wang:2014} & 86.3 \\
\newcite{wang:2015} & 86.6 \\
\hline
\end{tabular}
\end{center}
\caption{\label{final_ch} Final results on test set of CTB.}
\end{table}
\section{Analysis}
We analyze the results of section 23 in WSJ given by our model (i.e. in-order parser) and two baseline models (i.e. the bottom-up parser and the top-down parser) against the sentence length, the span length and the constituent type, respectively.
\begin{table}[!tp]
\begin{center}
\renewcommand{\arraystretch}{0.8}
\begin{tabular}{>{\small}l|>{\small}c>{\small}c}
\hline
Model &  UAS  & LAS \\
\hline
\hline
\newcite{dyer:2016} & 85.5 & 84.0 \\
\newcite{ballesteros:2016} & 87.7 & 86.2 \\
\newcite{kiperwasser:2016} & 87.6 & 86.1 \\
\newcite{cheng:2016} $\dagger$ & 88.1 & 85.7 \\
\newcite{dozat:2017} $\dagger$ & 89.3 & 88.2 \\
\hdashline
In-order parser & 87.4 & 86.4 \\
In-order parser -re & \textbf{89.4} & \textbf{88.4} \\
\hline
\end{tabular}
\end{center}
\caption{\label{final_dep_ch}  Dependency accuracy (\%) on CTB test set. $\dagger$ means graph-based parsing. ``-re" means supervised reranking.}
\end{table}
\subsection{Influence of sentence length}
\begin{figure}[!tp]
\begin{tikzpicture}[scale=1, font=\small]
\begin{axis} [
height = 5cm,
width = 8cm,
xlabel = Sentence length,
ylabel = F$_1$ (\%),
xmin=10,
xmax=60,
ymin=86,
ymax=95,
ytick pos=left,
legend style={at={(0.55,0.5)}},
ymajorgrids=true,
grid style=dashed,
y label style={at={(0.05,0.5)}}
]
\addplot[thick] coordinates{
	(10, 94.45)
	(20, 92.81)
	(30, 91.18)
	(40, 90.62)
	(50, 88.94)
	(60, 86.04)
};
\addlegendentry{Top-down parser}
\addplot[dashed] coordinates{
	(10, 94.59)
	(20, 92.94)
	(30, 91.84)
	(40, 90.49)
	(50, 88.44)
	(60, 86.23)
};
\addlegendentry{Bottom-up parser}
\addplot[thick,dash dot] coordinates{
	(10, 94.58)
	(20, 92.96)
	(30, 91.87)
	(40, 90.58)
	(50, 88.96)
	(60, 86.68)
};
\addlegendentry{In-order parser}
\end{axis}
\end{tikzpicture}
\caption{F$_1$ score against sentence length. (the number of words in a sentence, in bins of size 10, where 20 contains sentences with lengths in [10, 20).)}
\label{length} 
\end{figure}
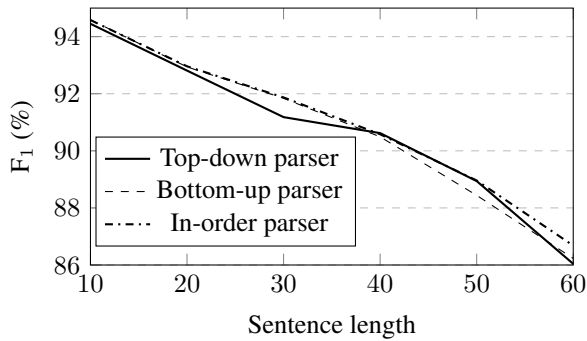

Figure \ref{length} shows the F$_1$ scores of the three parsers on sentences of different lengths.
Compared to the top-down parser, the bottom-up parser perform better on the short sentences with the length falling in the range [20-40].
This is likely because the bottom-up parser takes advantages of rich local features from partially-built trees, which are useful for parsing short sentences.
But these local structures are can be insufficient for parsing long sentences due to error propagation.
On the other hand, the top-down parser obtain better results on long sentences with the length falling in the range [40-50].
This is because, as the length of sentences increase, lookahead features become rich and they could be correctly represented by the LSTM, which is beneficial for parsing non-local structures.
We find that the in-order parser performs the best for both short and long sentences, showing the advantages of integrating bottom-up and top-down information.

\begin{table*}[!tp]
\begin{center}
\renewcommand{\arraystretch}{0.8}
\begin{tabular}{>{\small}l>{\small}l>{\small}c>{\small}c>{\small}c>{\small}c>{\small}c>{\small}c>{\small}c>{\small}c>{\small}c}
\hline
\multicolumn{2}{>{\small}l}{} & NP & VP & S & PP & SBAR & ADVP & ADJP & WHNP & QP \\
\hline
\hline
\multicolumn{2}{>{\small}l}{Top-down parser} & 92.87  & 92.51 & 91.36 & 87.96 & 86.74  & 85.21 & 75.41 & 96.44 & 89.41 \\
\multicolumn{2}{>{\small}l}{Bottom-up parser} & 93.01 & 92.20 & 91.46 & 87.95 & 86.81 & 84.58 & 74.84 & 94.99 &  89.95 \\
\hline
\multicolumn{2}{>{\small}l}{In-order parser} & 93.23 & 92.83 & 91.87 & 88.97 & 88.05 & 86.30 & 76.62 & 96.75 & 92.16 \\
\multicolumn{2}{>{\small}l}{Improvement} & +0.22 & +0.32 & +0.41 & +1.01 & +1.04 & +1.09 & +1.21 & +0.31 & +2.01 \\
\hline
\end{tabular}
\end{center}
\caption{\label{errors analysis on phrase types} Comparison on different phrases types.}
\end{table*}

\subsection{Influence of span length}
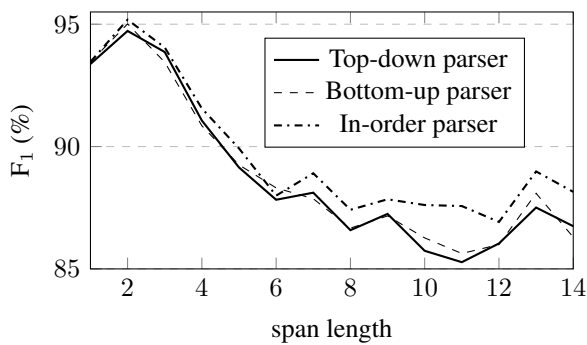
\begin{figure}[!tp]
\begin{tikzpicture}[scale=1, font=\small]
\begin{axis} [
height = 5cm,
width = 8cm,
xlabel = span length,
ylabel = F$_1$ (\%),
xmin=1,
xmax=14,
ymin=85,
ymax=95.5,
ytick pos=left,
legend style={at={(0.9,0.9)}},
ymajorgrids=true,
grid style=dashed,
y label style={at={(0.05,0.5)}}
]
\addplot[thick] coordinates{
	( 1 , 93.3742802303 )
( 2 , 94.7249983454 )
( 3 , 93.863008818 )
( 4 , 91.0624910625 )
( 5 , 89.1454105381 )
( 6 , 87.8260869565 )
( 7 , 88.1146389435 )
( 8 , 86.5833051707 )
( 9 , 87.2468117029 )
( 10 , 85.7394366197 )
( 11 , 85.2751611304 )
( 12 , 86.0401955459 )
( 13 , 87.5074716079 )
( 14 , 86.7452135493 )
};
\addlegendentry{Top-down parser}
\addplot[dashed] coordinates{
	( 1 , 93.4612432847 )
( 2 , 95.0115321252 )
( 3 , 93.4486139585 )
( 4 , 90.8492993995 )
( 5 , 89.250327164 )
( 6 , 88.3105022831 )
( 7 , 87.86775007 )
( 8 , 86.6666666667 )
( 9 , 87.1641791045 )
( 10 , 86.2676056338 )
( 11 , 85.6293359762 )
( 12 , 85.9956236324 )
( 13 , 88.0866425993 )
( 14 , 86.2973760933 )
};
\addlegendentry{Bottom-up parser}
\addplot[thick,dash dot] coordinates{
	( 1 , 93.4638024616 )
( 2 , 95.1828665568 )
( 3 , 94.0485721911 )
( 4 , 91.5573302909 )
( 5 , 89.9236356863 )
( 6 , 87.9945429741 )
( 7 , 88.9074228524 )
( 8 , 87.4163319946 )
( 9 , 87.8448918717 )
( 10 , 87.610619469 )
( 11 , 87.5686470295 )
( 12 , 86.9184455391 )
( 13 , 88.9825406382 )
( 14 , 88.1481481481 )
};
\addlegendentry{In-order parser}
\end{axis}
\end{tikzpicture}
\caption{F$_1$ score against span length.}
\label{span} 
\end{figure}

Figure \ref{span} shows the F$_1$ scores of the three parsers on spans of different lengths.
The trend of performances of the two baseline parsers are similar.
Compared to the baseline parsers, the in-order parser obtains significant improvement on long spans.
It is linguistically because the in-order traversal over a tree allows constituent types of spans to be correctly projected based on the information of the beginning (leftmost nodes) of the spans, and then the projected constituents constrain long span construction, which is different from the top-down parser, which generates constituent types of spans without trace of the spans.
\subsection{Influence of constituent type}
Table \ref{span} shows the F$_1$ scores of the three parsers on frequent constituent types.
The bottom-up parser performs better than the top-down parser on constituent types including NP, S, SBAR, QP.
We find that the prediction of these constituent types requires explicitly modeling of bottom-up structures.
In other words, bottom-up information is necessary for us to know if the span can be a noun phrase (NP) or sentence (S) for example.
On the other hand, the top-down parser has better performance on WHNP, which can be due to the reason that a WHNP starts with a certain question word, which makes the prediction is easy without bottom-up information.
The in-order parser performs the best on all constituent types, which demonstrates that the in-order parser can benefit from both bottom-up and top-down information. 

\subsection{Examples}
We give output examples from the test set to qualitatively compare the performances of the three parsers using the fully-supervised model without reranking, as shown in Table \ref{example}.
For example, given the sentence \#2006, the bottom-up and the in-order parsers give both correct results.
However, the top-down parser makes an incorrect decision to generate an S, which leads subsequent incorrect decisions on VP to complete S. 
Sentence pattern ambiguaty allows top-down guidance to over-parsing the sentence by recognizing the word ``Plans" as a verb, while more bottom-up information is useful for the local disambiguation.
\begin{table*}[!tp]
\begin{center}
\renewcommand{\arraystretch}{0.8}
\begin{tabular}{>{\small}l|>{\small}l}
\hline
Sent \#2066 & Employee Benefit Plans Inc. -- \\
\hline
Gold & (NP Employee Benefit Plans Inc. -) \\
Top-down & \textcolor{red}{(S }(NP Employee Benefit \textcolor{red}{)} \textcolor{red}{(VP} Plans \textcolor{red}{(NP} Inc. \textcolor{red}{)} -- \textcolor{red}{)} )\\
Bottom-up & (NP Employee Benefit Plans Inc. -) \\
In-order  & (NP Employee Benefit Plans Inc. -) \\
\hline
Sent \#308 & ... whether the new posted prices will stick once producers and customers start to haggle .\\
\hline
Gold & ... (VP will (VP stick (SBAR once (S (NP producers and customers ) (VP start (S ...) ) ) ) ) ) ... \\
Top-down & ... (VP will (VP stick (SBAR once (S (NP producers and customers ) (VP start (S ...) ) ) ) ) ) ... \\
Bottom-up &  ... (VP will (VP stick \textcolor{red}{(NP} once producers and customers \textcolor{red}{)} ) ) ... (VP start (S ...) ) ... \\
In-order & ... (VP will (VP stick (SBAR once (S (NP producers and customers ) (VP start (S ...) ) ) ) ) ) ... \\
\hline
Sent \#1715 & This has both made investors uneasy and the corporations more vulnerable .\\
\hline
Gold & (S (NP This) (VP has (VP both made (S (S investors uneasy) and (S the corporations ...)))) .) \\
Top-down &  (S \textcolor{red}{(S} (NP This) (VP has (S (NP both) (VP made investors uneasy)))\textcolor{red}{)} and (S the corporations ...) .) \\
Bottom-up & (S \textcolor{red}{(S} (NP This) (VP has (S both (VP made investors uneasy)))\textcolor{red}{)} and (S the corporations ...) .) \\
In-order & (S (NP This) (VP has both \textcolor{red}{(VP} made (S (S investors uneasy) and (S the corporations ...))\textcolor{red}{)}) .) \\
\hline
\end{tabular}
\end{center}
\caption{\label{example} Output examples of the three parsers on the English test set. Incorrect constituents are marked in red.}
\end{table*}


Given the sentence \#308, the bottom-up parser prefers construction of local constituents such as ``once producers and customers", ignoring the possible clause SBAR, however, which is captured by the in-order parser, because the parser projects a constituent SBAR from the word ``stick" and continues to complete the clause, showing that top-down lookahead information is necessary for non-local disambiguation.
The in-order parser gives the correct output for the sentence \#2066 and the sentence \#308, showing that it can benefit from bottom-up and top-down information.

In the sentence \#1715, there are coordinated objects such as ``investors uneasy" and ``the corporations more vulnerable".
All of the three parsers can recognize coordination.
However, the top-down and the bottom-up parsers incorrectly recognize the ``This has both made investors uneasy" as a complete sentence.
The top-down parser incorrectly generates S, marked in red, at a early stage, leaving no choice but to follow this incorrect non-terminal.
The bottom-up parser without lookahead information makes incorrect local decisions.
In contrast, the in-order parser reads the word ``and" and projects a non-terminal S for coordination after completing ``(S investors uneasy)".
On the other hand, the in-order parser is confused to project for the word ``made" or the word ``both" into a VP, which we think could be addressed by using a in-order system variant with $k$=2 described in section 3.
\section{Related work}
Our work is related to left corner parsing.
\newcite{rosenkrantz:1970} formalize this in automata theory, which have appeared frequently in the compiler literature.
\newcite{roark:1999} apply the strategy into parsing. 
Typical works investigate the transformation of syntactic trees based on left-corner rules \cite{roark:2001,schuler:2010,van:2013}.
In contrast, we propose a novel general transition-based in-order constituent parsing system.

Neural networks have achieved the state-of-the-art for parsing under various grammar formalisms, including dependency \cite{dozat:2017} constituent \cite{dyer:2016,kuncoro:2017} and CCG parsing \cite{xu:2016,lewis:2016}.
Seminal work employs transition-based methods \cite{chen:2014}.
This method has been extended by investigating more complex representations of configurations for constituent parsing \cite{watanabe:2015,dyer:2016}.
\newcite{dyer:2016} employ stack-LSTM onto top-down system, which is the same as our top-down parser.
\newcite{watanabe:2015} employ tree-LSTM to model the complex representation in stack in bottom-up system.
We are the first to investigate in-order traversal by designing a novel transition-based system under the same neural structure model framework.

\section{Conclusion}
We proposed a novel psycho-linguistically motivated constituent parsing system based on the in-order traversal over syntactic trees, aiming to find a compromise between bottom-up constituent information and top-down lookahead information. 
On standard WSJ benchmark, our in-order system outperforms bottom-up parsing on non-local ambiguity and top-down parsing on local decision.
The resulting parser achieves the state-of-the-art constituent parsing results by obtaining 94.2 F$_1$ and dependency parsing results by obtaining 96.2\% UAS and 95.2\% LAS.

\section*{Acknowledgments}
We thank the anonymous reviewers for their detailed and constructive comments.
Yue Zhang is the corresponding author.

\bibliography{tacl}
\bibliographystyle{acl2012}

\end{document}